\pdfoutput=1

\documentclass[11pt]{article}

\usepackage[]{acl}
\usepackage{times}
\usepackage{latexsym}
\usepackage{graphicx}

\usepackage[T1]{fontenc}

\usepackage[utf8]{inputenc}

\usepackage{microtype}

\usepackage{newfloat}
\usepackage{listings}
\usepackage{threeparttable}
\usepackage{booktabs}
\usepackage{ulem}
\usepackage{algorithm}
\usepackage{algorithmic}
\usepackage{amsmath}
\usepackage{amsfonts}
\usepackage{multirow}
\usepackage{caption}
\usepackage{makecell}

\newcommand{\eat}[1]{{}}

\title{Can Pre-trained Language Models Interpret Similes as \underline{Smart} as Human?}

\author{Qianyu He\textsuperscript{\rm 1}\thanks{\ \ Equal contribution}, \quad
        Sijie Cheng\textsuperscript{\rm 1}\footnotemark[1],\quad
        Zhixu Li\textsuperscript{\rm 1}\thanks{\ \ Corresponding author},\quad
        Rui Xie\textsuperscript{\rm 3},\quad
        Yanghua Xiao\textsuperscript{\rm 1,2}\footnotemark[2]\quad \\
        
        \textsuperscript{\rm 1}Shanghai Key Laboratory of Data Science, School of Computer Science, Fudan University \\ 

         \textsuperscript{\rm 2}Fudan-Aishu Cognitive Intelligence Joint Research Center, Shanghai, China \\
        \textsuperscript{\rm 3}Meituan, Shanghai, China \\
        \texttt{qyhe21@m.fudan.edu.cn}, \texttt{rui.xie@meituan.com}, \\
        \texttt{\{sjcheng20, zhixuli, shawyh\}@fudan.edu.cn}

        }
        
\begin{document}
\maketitle
\begin{abstract}
Simile interpretation is a crucial task in natural language processing.
Nowadays, pre-trained language models (PLMs) have achieved state-of-the-art performance on many tasks.
However, it remains under-explored whether PLMs can interpret similes or not.
In this paper, we investigate the ability of PLMs in simile interpretation by designing a novel task named Simile Property Probing, i.e., to let the PLMs infer the shared properties of similes.
We construct our simile property probing datasets from both general textual corpora and human-designed questions, containing 1,633 examples covering seven main categories.
Our empirical study based on the constructed datasets shows that PLMs can infer similes' shared properties while still underperforming humans.
To bridge the gap with human performance, we additionally design a knowledge-enhanced training objective by incorporating the simile knowledge into PLMs via knowledge embedding methods.
Our method results in a gain of 8.58\% in the probing task and 1.37\% in the downstream task of sentiment classification.
The datasets and code are publicly available at \href{https://github.com/Abbey4799/PLMs-Interpret-Simile}{https://github.com/Abbey4799/PLMs-Interpret-Simile}. 

%


\end{abstract}

\section{Introduction}

\begin{figure}[t]
    \centering
        \includegraphics[width=1\linewidth]{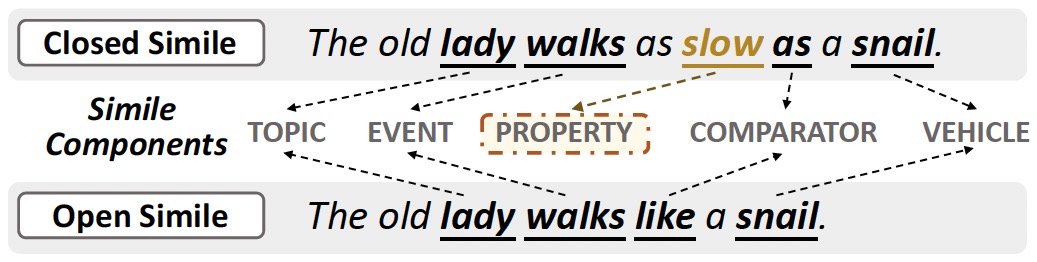} 
    \captionsetup{font={small}} 
    \caption{Examples of two types of similes. Whether the component \textit{property} is stated determines the type of simile.}
    \label{fig:two_types_simile}
\end{figure}

A simile is a figure of speech comparing two fundamentally different entities via shared properties \cite{paul1970figurative}.
There are two types of similes as illustrated in Figure~\ref{fig:two_types_simile}, \textit{closed} similes explicitly reveal the {\em shared properties} between the {\em topic entity} and the {\em vehicle entity}, such as the property \textit{``slow''} shared by \textit{``lady''} and \textit{``snail''} in the sentence \textit{``The old lady walks as slow as a snail''}; while \textit{open} similes do not state the shared property such as the sentence \textit{``The old lady walks like a snail''}.
Similes play a vital role in human expression to make literal utterances more vivid and graspable and are widely used in the corpus of various domains \cite{liu2018neural, chakrabarty-etal-2020-generating,zhang2020writing}. It is estimated that over 30\% of the comparisons can be regarded as similes in product reviews~\cite{niculae2014brighter}.

\begin{table*}[t] 
\small
    \centering
        \resizebox{\textwidth}{24mm}{
        \begin{tabular}{ccc}
        \toprule  
        \textbf{Category} & \textbf{Question Example} & \% \\

        \midrule  
        \textbf{Qualities} & 
        My \textit{client} is as \texttt{[MASK]} as a newborn \textit{lamb} . \underline{\textbf{A.} innocent} \textbf{B.} delicious \textbf{C.} legal  \textbf{D.} guilty & 27.78 \\
        \midrule
        \textbf{Condition} & The \textit{toddler} was running around as \texttt{[MASK]} as a \textit{bee}. \underline{\textbf{A.} busy} \textbf{B.} yellow \textbf{C.} idle \textbf{D.} messy & 22.28 \\
        \midrule 
        \textbf{Sense} & His \textit{anger} was as \texttt{[MASK]} as a burning \textit{ember}. \underline{\textbf{A.} hot} \textbf{B.} red \textbf{C.} cold \textbf{D.} warm  & 17.20 \\
        \midrule
        \textbf{Measurement} & My new baby \textit{brother} is as \texttt{[MASK]} as a \textit{button}. \textbf{A.} red \underline{\textbf{B.} tiny}  \textbf{C.} cute \textbf{D.} hot  & 14.16 \\
        \midrule
        \textbf{Color} & He was scared so much. \textit{He} was as \texttt{[MASK]} as a \textit{ghost}. \underline{\textbf{A.} white } \textbf{B.} holy \textbf{C.} gay \textbf{D.} black & $\ \ $06.75 \\
        \midrule
        \textbf{Time} &	The old \textit{man} walks as \texttt{[MASK]} as a \textit{tortoise}. \textbf{A.} young \textbf{B.} little \underline{\textbf{C.} slow }  \textbf{D.} quick & $\ \ $06.57 \\
        \midrule
        \textbf{Emotion} & The \textit{boy} was as \texttt{[MASK]} as a \textit{dog} that lost its bone. \textbf{A.} happy \textbf{B.} friendly \underline{\textbf{C.} sad} \textbf{D.} glad & $\ \ $05.26 \\

        \bottomrule 
        \end{tabular}}
        \caption{Percentage and examples for our simile probes of different categories. The option marked with ``$\underline{\ \ \ \ \ \ }$'' indicates the correct answer. The italicized words one by one in each sentence are the topic, masked property, and vehicle, respectively. }
    \label{tab:category_statistics}
\end{table*}


Simile interpretation is a crucial task in natural language processing~\cite{veale2007learning,qadir2016automatically,chakrabarty2021s}, which can assist several downstream tasks such as understanding more sophisticated figurative language~\cite{veale2007learning} and sentiment analysis~\cite{niculae2014brighter,qadir2015learning}. 
Take the simile \textit{``the lawyer is like a shark''} for an example. 
Although all words in this simile are neutral, this simile expresses a negative affect since \textit{``lawyer''} and \textit{``shark''} share the negative property \textit{``aggressive''}.

\eat{While it is usually easy to identify the shared properties of closed similes through synthetic patterns like {\it ``as ...as''}~\cite{veale2007learning,veale2007comprehending}, the critical issue in similes understanding lies in inferring the shared properties of similes.
It is nontrivial given that the two objects being compared are fundamentally different types of entities, such as \textit{``lady''} and \textit{``snail''} in the previous example.
It may require our deep understanding of the two entities, such as having the commonsense knowledge that \textit{``old people always usually walk slowly''} and \textit{``snails move at a slow speed''}, to reveal their shared property.
}

In the past few years, large pre-trained language models (PLMs) have achieved state-of-the-art performance on many natural language processing tasks \cite{devlin2018bert,liu2019roberta}.
Recent studies suggest that PLMs have possessed various kinds of knowledge into contextual representations \cite{goldberg2019assessing,petroni2019language,lin2019kagnet,cui2021commonsense}.
%
However, the ability of PLMs to interpret similes remains under-explored.
Although some recent work~\cite{chakrabarty2021s} studies the ability of PLMs in choosing or generating the plausible continuations in narratives, this way cannot fully reveal the ability of PLMs to interpret similes.


In this paper, we propose to investigate the ability of PLMs in simile interpretation by designing a novel task named as \textit{Simile Property Probing}, i.e., to let the PLMs infer the shared properties of similes.
%
Specifically, we design a particular masked-word-prediction probing task in the form of multiple-choice questions. This probe masks the explicit property of a \textit{closed} simile and then lets the PLMs discriminate it from three distractors.
To make the questions convincing and challenging, the distractors should be not only \textit{true-negative} as they would introduce logical errors once they are filled in the sentence, but also \textit{challenging} as they are semantically close to the correct answer.
To achieve this, we propose to obtain some similar properties of the golden one from ConceptNet~\cite{liu2004conceptnet} and COMET~\cite{bosselut2019comet}, from which we select the three best distractors according to their proximity to the golden property in the feature space.
%
From two different types of data sources: textual corpus collection and human-designed questions, we collect a total of 1,633 probes with various usage frequencies and context diversities, covering seven categories as listed in Table~\ref{tab:category_statistics}.
%
%
%
%
%
%
%
%

Based on our designed task, we evaluate the ability of BERT \cite{devlin2018bert} and RoBERTa \cite{liu2019roberta} to infer the shared properties of similes.
We perform an empirical evaluation in two settings: (1) zero-shot, where the models are off-the-shelf; (2) fine-tuned, where the models are fine-tuned with MLM objective via masking properties. 
We observe that PLMs have been able to infer properties of similes in the pre-training stage and the ability can be further enhanced by fine-tuning.
However, fine-tuned PLMs still perform worse than humans.
Moreover, we find that the simile components \textit{vehicle} and \textit{topic} contribute the most when inferring the properties.

Inspired by the sufficient hints offered by the components \textit{vehicle} and \textit{topic} in our empirical study, we propose a knowledge-enhanced training objective to further bridge the gap with human performance.
Considering \textit{property} (p) as the \textbf{relation} between \textit{topic} (t) and \textit{vehicle} (v), we design a simile knowledge embedding objective function following conventional knowledge embedding methods \cite{bordes2013translating} to incorporate the simile knowledge \textit{(t,p,v)} into PLMs.
To integrate simile knowledge and language understanding into PLMs, we jointly optimize the knowledge embedding objective and the MLM objective in our design.
Overall, the knowledge-enhanced objective shows effectiveness in our probing task and the downstream task of sentiment classification.


\begin{figure*}[!t]
    \centering
    \vspace{-0.4cm}
        \includegraphics[width=1\linewidth]{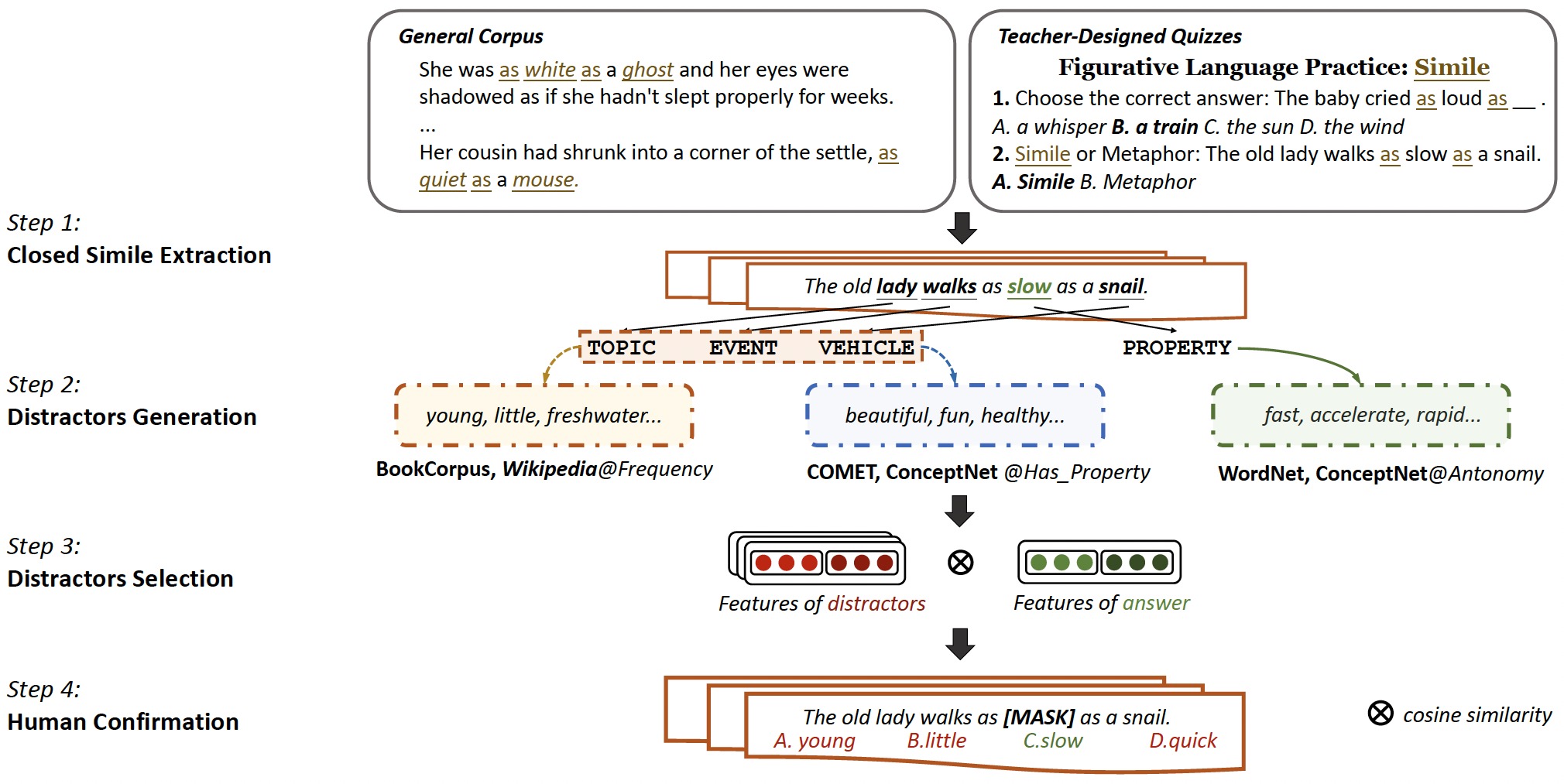} 
    \vspace{-0.3cm}
    \captionsetup{font={small}} 
    \caption{A process for designing our simile property probing task. In Step 1, we collect closed similes from two different sources. In Step 2, according to four important components in each simile, we generate distractor candidates with three strategies. In Step 3, we adopt cosine similarity to select more challenging distractors. In Step 4, we ask human annotators to ensure the quality and obtain our final probing datasets.}
    \label{fig:framework}
\end{figure*}

To summarize, our contributions are three-fold:
(1) To our best knowledge, we are the first to systematically evaluate the ability of PLMs in interpreting similes via a proposed novel simile property probing task.
(2) We construct simile property probing datasets from both general textual corpora and human-designed questions, and the probing datasets contain 1,633 examples covering seven main categories of similes.
(3) We also propose a novel knowledge-enhanced training objective by complementing the MLM objective with the knowledge embedding objective. This method gains 8.58\% in the probing task and 1.37\% in the downstream task of sentiment classification.

\section{Preliminaries on Simile}
A sentence of simile generally consists of five major components \cite{hanks2013lexical,niculae2014brighter}, where four are necessary and the remaining one is optional. 
The four explicit components are as follows:
(1) \textbf{topic (or tenor)}: the subject of the comparison acting as source domain;
(2) \textbf{vehicle}: the object of the comparison acting as target domain;
(3) \textbf{event}: the predicate indicating act or state;
(4) \textbf{comparator}: the trigger word of a simile such as \textit{as} or \textit{like}.
The optional component \textbf{property} reveals the shared characteristics between the topic and the vehicle.
There are two types of similes depending on whether the property is explicit or implicit \cite{beardsley1981aesthetics}.
The similes which mention the property directly are named as the \textit{closed} similes, while the others are \textit{open} similes, as shown in Figure \ref{fig:two_types_simile}.

\section{The Simile Property Probing Task}

\subsection{Task Formulation}

%
%

To estimate the ability of PLMs in simile interpretation, we design a particular {\it Simile Property Probing} task, which masks the explicit property of a \textit{closed} simile, and then lets the PLMs discriminate it among four candidates.
Considering that the shared properties between topic and vehicle may not be unique~\cite{lacroix2005interpretive}, we specifically design a multiple-choice question answering task (with only one correct answer) rather than a cloze task to probe the ability of PLMs to infer properties of similes, since the latter one may result in multiple correct answers.

Formally, given a simile text sequence $S = ( w_1, w_2, ..., w_{i-1}, \texttt{[MASK]}, w_{i+1}, ...,w_N )$, where the shared property $w_i$ between the topic and vehicle is masked, the probing task requires the PLMs to find the correct property from four options, where the other three options are hard distractors.



    


\subsection{Probing Data Collection}
We construct datasets for the proposed probing task in four steps. 
The overview of our probing data collection process is described in Figure \ref{fig:framework}. 

%


\subsubsection{Data Sources} \label{sec:data_sources}
We construct our datasets from two different sources to detect the capability of PLMs from two perspectives: textual corpus collection and human-designed questions.
%
To avoid laborious human labeling on the implicit properties of open similes, we collect closed similes with explicit properties.

\textbf{General Corpus.} Following \cite{hanks2005similes,niculae2013computational}, we adopt two general corpora, British National Corpus (BNC)$\footnote{https://www.english-corpora.org/bnc/}$ and iWeb$\footnote{https://www.english-corpora.org/iweb/}$.
To identify closed similes, we extract the sentences matching the syntax \textit{as ADJ as (a, an, the) NOUN}. 
Through syntactic pattern matching, we finally collect 1,917 sentences.

\textbf{Teacher-Designed Quizzes.} Questions about similes designed by teachers from educational resources are ideal sources for assessing the ability to understand similes.
Hence, we choose Quizizz$\footnote{https://quizizz.com/}$, an emerging learning platform founded in 2015.
On this platform, users can create quizzes on a specific topic as teachers to assess students' understanding of related knowledge.
We collect a set of quizzes with titles concerning similes and extract the complete closed simile sentences from the questions and answers in these quizzes.
Finally, we retrieve 875 complete closed similes from 1,235 quizzes.

To assure the quality of our constructed datasets and prepare for further analysis, three annotators are required to decide whether the extracted sentences are similes or not, and annotate their corresponding simile components.
The inter-annotator agreement on identifying similes is 0.77 using Fleiss' Kappa score \cite{fleiss1971measuring}.
All the properties in our datasets are single-token by replacing multi-token properties with their single-token synonyms in the knowledge base WordNet \cite{miller1995wordnet} and ConceptNet \cite{liu2004conceptnet}.

\begin{figure}[t]
    \centering
        \vspace{-0.4cm}
        \includegraphics[width=0.9\linewidth]{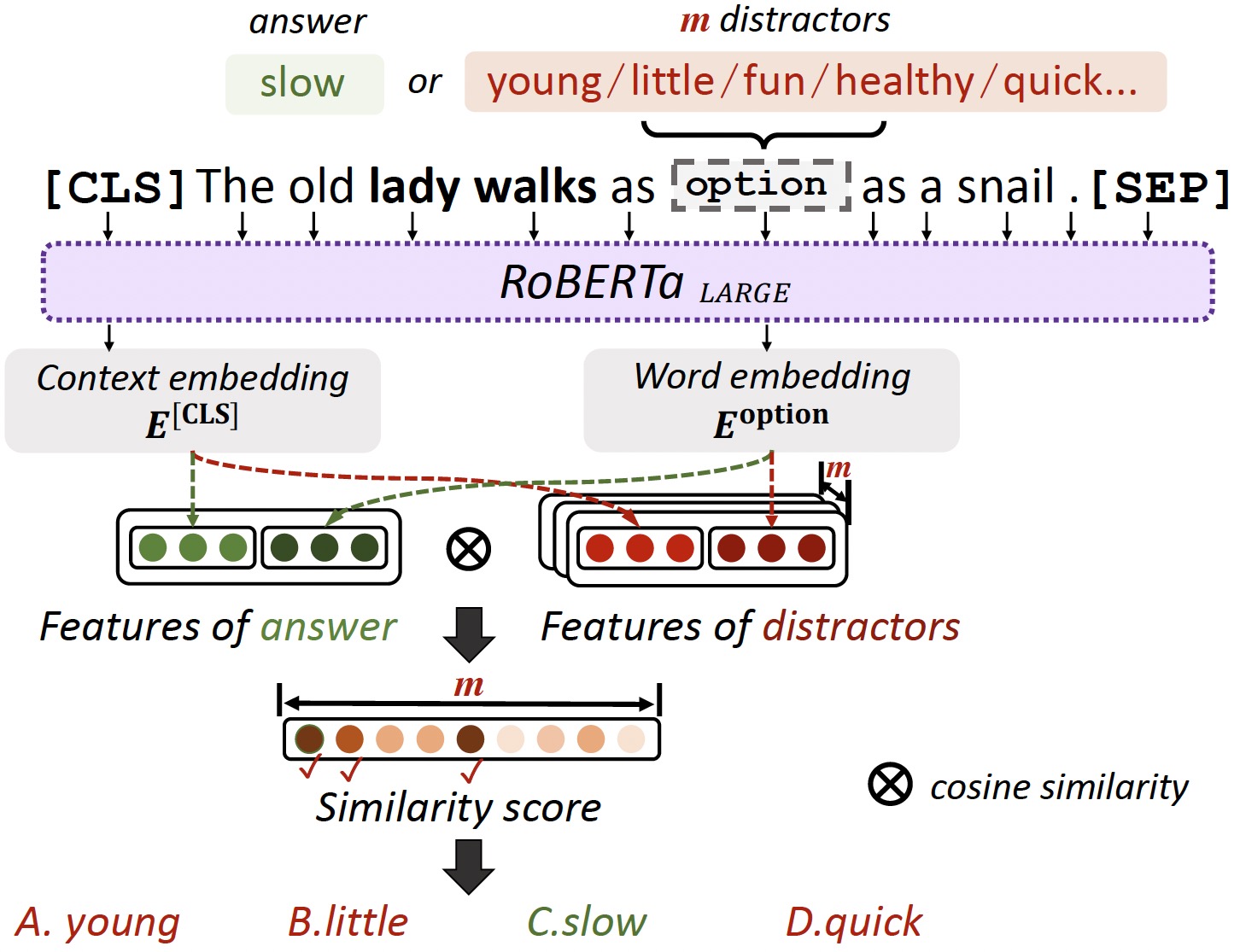} 
        \vspace{-0.1cm}
    \captionsetup{font={small}} 
    \caption{Illustration of the distractor selection method.}
    \vspace{-0.3cm}
    \label{fig:similarity_framework}
\end{figure}

\subsubsection{Distractor Design} \label{sec:distractor_design}

To make our probes convincing, three distractors are designed against the original property in each simile with two criteria \cite{haladyna2002review,ren2020knowledge}: \textit{true-negative} and \textit{challenging}.
%
%
We argue that well-designed distractors should be illogical when filled into the questions \textit{(true-negative)} while being semantically related to the correct answer \textit{(challenging)}.
Our distractor design mainly involves three phases: 1) distractor generation; 2) distractor selection; 3) Human Confirmation.

\textbf{Distractor Generation.}
To meet the requirement of \textit{challenging}, we generate distractor candidates from the four semantic-related components of a simile, i.e., topic, vehicle, event, and property.
Given the original property, we harvest its antonyms from the knowledge base WordNet and ConceptNet.
With regard to three other components, we extract their properties from two sources as follows.
Given a component, we utilize the \textit{HasProperty} relation from ConceptNet \cite{liu2004conceptnet} and COMET \cite{bosselut2019comet} to retrieve the property. 
Moreover, we rank the adjectives or adverbs concerning\footnote{We adopt dependency parsing via the StanfordNLP tool to find adjectives and adverbs related to components.} each component in Wikipedia and BookCorpus corpus\footnote{https://huggingface.co/datasets/} by frequency and select the top ten candidates with a frequency of more than one.

\begin{table}[t] 
    \scriptsize
    \vspace{-0.5cm}
    \resizebox{0.48\textwidth}{30mm}{
    \centering
        \begin{tabular}{cccc}
        \toprule  
        \textbf{Dataset} & \begin{tabular}[c]{@{}l@{}}\textbf{General}\\\textbf{Corpus}\end{tabular} & \textbf{Quizzes} \\
        \midrule  
        \textbf{\#Sentence} & 775 & 858 \\ 
        \midrule 
        \textbf{\#Unique \textit{topic} concept} & 415 & 366 \\
        \textbf{\#Unique \textit{property} concept} & 280 & 160 \\
        \textbf{\#Unique \textit{vehicle} concept} & 522 & 250\\
        \textbf{\#Unique \textit{event} concept} & 147 & 66 \\
        \textbf{\#Unique \textit{topic-vehicle} pair} & 743 & 684 \\
        \textbf{\#Unique \textit{topic-property-vehicle} pair} &  751 & 701 \\
        \midrule 
        \textbf{Maximum sentence length} & 98 & 44 \\
        \textbf{Average sentence length} & 25.80 & 12.69\\
        \textbf{Minimum sentence length} & 7 & 7\\
        \midrule 
        \textbf{@Start} & 34.32\% & 20.40\% \\
        \textbf{@Middle} & 43.23\% & 63.29\% \\
        \textbf{@End} & 22.45\% & 16.32\% \\
        \bottomrule 
        \end{tabular}}
        \captionsetup{font={small}} 
        \caption{Statistics of our simile property probing datasets. @ denotes the position of the simile in the given sentence.}
        
    \label{tab:corpus_statistics}
\end{table}

\textbf{Distractor Selection.}
To select the most \textit{challenging} distractors from the generated distractor candidates, we propose to measure the similarity between the original sentence with the correct property and the sentence with a distractor.
Intuitively, the more similar the two sentences, the more challenging the distractor.
An example of the distractor selection process is depicted in Figure \ref{fig:similarity_framework}.
Given the original sentence or the new sentence replacing the correct property with a distractor, we first utilize $\text{RoBERTa}_{\text{LARGE}}$ to extract two types of features.
One feature is context embedding, which is the sentence embedding of \texttt{[CLS]}, while the other feature is word embedding, which is the token embedding of the answer or distractors. 
We then concatenate the embeddings of the two features to compute the cosine similarity between sentences with the answer and a distractor.
Finally, we select the top 3 distractors with the highest similarities.

\textbf{Human Confirmation.}  
To ensure the distractors are {\it true-negative}, three human annotators are asked to label each selected distractor. If more than two annotators are uncertain about its correctness, we replace it with another suitable candidate.


\subsubsection{Statistics of the Datasets} \label{sec:statistics_of_the_datasets}
Table \ref{tab:corpus_statistics} presents the statistics of our constructed datasets.
We count unique components and component pairs to present the usage frequencies of similes.
The length of the sentences in each dataset indicates the diversities of context.
Additionally, we analyze the distribution of the position of simile in the sentences in each dataset, where \textit{start}, \textit{middle} and \textit{end} correspond to the positions of the three equally divided parts of each sentence.
We also investigate the categories covered by our datasets.
The results and details about the category classification are provided in Appendix \ref{sec:dataset_description_appendix}.
Overall, the Quizzes dataset provides similes commonly expressed by people, while the General Corpus dataset presents similes with more diverse contexts.

\subsection{Supervision for Fine-Tuning PLMs} \label{sec:supervised}
Besides evaluating the ability of PLMs in the zero-shot setting where the models are off-the-shelf, we also study whether the performance could be improved through fine-tuning with the MLM objective via masking properties.
To achieve this, we collect training data from Standardized Project Gutenberg Corpus$\footnote{https://github.com/pgcorpus/gutenberg/}$ (SPGC) \cite{gerlach2020standardized}.
SPGC is a 3 billion words corpus collected from about 60 thousand eBooks.
We extract similes via matching the syntactic pattern \textit{(Noun ... as ADJ as ... NOUN)} and end up with 4,510 sentences. 
Additionally, we adopt dependency parsing$\footnote{https://stanfordnlp.github.io/CoreNLP/}$ to automatically annotate the simile components of each sentence without human labor.

\section{Empirical Study on PLMs}
In this section, we first conduct a set of experiments to probe the ability of PLMs to infer properties in similes and then evaluate the influence of each component on the model performance.

\subsection{Ability to Infer Shared Properties} \label{sec:ability_to_infer}
\subsubsection{Experiment Set-up}
To disentangle what is captured by the original representations and what is introduced from fine-tuning stage, we apply two different types of settings: (1) zero-shot; (2) fine-tuning.
In our first setting, we use BERT and RoBERTa with pre-trained masked-word-prediction heads to perform our probing task. 
In the second setting, we utilize the MLM training objective inherited from PLMs to fine-tune the models.
We replace the property of each simile with the special token \texttt{[MASK]} in our constructed supervised datasets (Section \ref{sec:supervised}) and ask models to recover the original property.
The experimental details are provided in the Appendix \ref{sec:experimental_details_appendix}.


We mainly compare the model accuracy of PLMs with the following baselines:
(1) \textbf{EMB} \cite{qadir2016automatically}: 
It obtains the composite simile vector by performing an element-wise sum of the word embedding for the vehicle and event, 
then selects the answer with the shortest cosine distance from the composite vector.
(2) \textbf{Meta4meaning} \cite{xiao2016meta4meaning} : 
This method prefers the properties which are strongly associated with both topic and vehicle.
It also prefers the properties that are more relevant to the vehicle than to the topic.
The association is measured by statistical significance.
(3) \textbf{ConScore} \cite{zheng2019love} :
It suggests that better properties would have a smaller and balanced distance to the topic and vehicle in the word embedding space.
(4) \textbf{MIUWE} \cite{bar2020automatic} :
The ranking method assigns each property a list of scores, including the statistical co-occurrences and similarity to the collocations of the topic and vehicle.
The baselines above mainly consider the statistical information and embedding similarities between the properties and the simile components.
The other baseline is human performance.
We sample 250 random questions from both datasets, and for each question, we gather answers from three people.
We take the majority vote as the human performance of our probing task and ensure that three annotators agree on the questions that they gave completely different annotation results.
%

\begin{table}[t] 
    \resizebox{0.48\textwidth}{24mm}{
    \centering
    \begin{tabular}{cccccc}
    \toprule
        \textbf{Setting} & \textbf{Models} & \begin{tabular}[c]{@{}l@{}}\textbf{General}\\ \textbf{Corpus}\end{tabular} & \textbf{Quizzes}  & \textbf{Gain} \\
        \midrule
         & \textbf{ConScore} \cite{zheng2019love} & 27.48 & 34.85 & - \\ 
         & \textbf{Meta4meaning} \cite{xiao2016meta4meaning}  & 27.74 & 47.44 & - \\ 
         & \textbf{EMB} \cite{qadir2016automatically} & 28.27 & 47.90  & - \\ 
         & \textbf{MIUWE} \cite{bar2020automatic}& 30.97 & 53.85 & - \\ 
        \midrule
        \multirow{4}{*}{\textbf{Zero-Shot}} & \textbf{$\text{BERT}_{\text{BASE}}$} & 64.13 & 74.36  & - \\
        & \textbf{$\text{BERT}_{\text{LARGE}}$} & 72.39 & 83.22  & - \\
        & \textbf{$\text{RoBERTa}_{\text{BASE}}$} & 69.55  & 82.87 & -  \\
        & \textbf{$\text{RoBERTa}_{\text{LARGE}}$} & 78.97 & 87.41  & - \\
        \midrule
        \multirow{4}{*}{\textbf{Fine-tuned}} & \textbf{MLM-$\text{BERT}_{\text{BASE}}$} &  67.74 & 82.05  &  +5.65\\
        & \textbf{MLM-$\text{BERT}_{\text{LARGE}}$}  & 73.85  & 84.58  &  +1.40\\
        & \textbf{MLM-$\text{RoBERTa}_{\text{BASE}}$}  & 70.58  & 84.69 & +1.43 \\
        & \textbf{MLM-$\text{RoBERTa}_{\text{LARGE}}$}  & \textbf{78.97} & \textbf{88.97} & +0.78 \\
        \midrule
        & \textbf{Human Performance} & 87.60 & 93.60  & - \\ 
    \bottomrule
    \end{tabular}}
        \captionsetup{font={small}} 
    \caption{Accuracy of different models in our simile property probing task.}
        \label{tab:probing_result}
\end{table}

\subsubsection{Results}

The accuracies of different methods under two different settings on our datasets are listed in Table~\ref{tab:probing_result}, where the last column represents the average absolute gains of each PLM after fine-tuning with the MLM objective.
All the results of our experiments are averaged over three random seeds.
%
First of all, the prediction accuracies of both BERT and RoBERTa in the zero-shot setting are much higher than the baselines only considering the statistical information and embedding similarities between simile components.
This phenomenon indicates that the knowledge learning from the pre-train stage can help infer the simile properties. 
Moreover, the performance can be further improved by training with the MLM objective, demonstrating that the fine-tuning phase with the supervised dataset can introduce related knowledge about similes.
However, models still underperform humans by several accuracy points, leaving room for improvement in our probing task.
%

Overall, all the models perform better on Quizzes Dataset than on General Corpus Dataset, indicating that more diverse contexts increase the difficulty of inferring the shared properties.
Also, RoBERTa consistently outperforms BERT, likely due to a larger pre-training corpus containing more similes.
More complementary results are provided in the Appendix \ref{sec:performance_on_different_categories_appendix}.

\begin{table*}[t]
\small
    \centering
    \vspace{-0.3cm}
    \begin{tabular}{cccccccc}
    \toprule
       \textbf{Datasets}  &    \textbf{Models} & \textbf{Topic} & \textbf{Vehicle} & \textbf{Event} & \textbf{Comparator} & \textbf{Random} \\
        \midrule
          \multirow{4}{*}{\begin{tabular}[c]{@{}l@{}}\textbf{General}\\ \textbf{Corpus}\end{tabular}} & 
            \textbf{$\text{BERT}_{\text{BASE}}$} & 59.87\textit{(-04.26)} & 54.58\textit{(-09.55)} & 62.84\textit{(-01.29)} & 46.32\textit{(-17.81)} & 63.05\textit{(-01.08)} \\
      &  \textbf{$\text{BERT}_{\text{LARGE}}$} & 67.74\textit{(-04.65)} & 61.16\textit{(-11.23)} & 70.19\textit{(-02.20)} &  46.06\textit{(-26.33)}& 69.07\textit{(-03.32)}\\
      &  \textbf{$\text{RoBERTa}_{\text{BASE}}$} & 65.29\textit{(-04.26)} &61.03\textit{(-08.52)} & 68.52\textit{(-01.03)} & 50.32\textit{(-19.23)}   & 67.31\textit{(-02.24)}\\
       & \textbf{$\text{RoBERTa}_{\text{LARGE}}$} & 76.90\textit{(-02.07)} & 69.68\textit{(-09.29)} & 77.55\textit{(-01.42)} & 54.97(-24.00) & 77.72\textit{(-01.25)}\\
       
        \midrule
         \multirow{4}{*}{\textbf{Quizzes}} & 
                 \textbf{$\text{BERT}_{\text{BASE}}$}  &  67.02\textit{(-07.34)} & 62.35\textit{(-12.01)} & 73.43\textit{(-00.93)} & 52.80\textit{(-21.56)}   & 71.91\textit{(-02.45)} \\
        &   \textbf{$\text{BERT}_{\text{LARGE}}$} &  77.86\textit{(-05.36)} & 64.57\textit{(-18.65)} & 82.63\textit{(-00.59)} & 55.24\textit{(-27.98)} & 79.91\textit{(-03.31)}\\
      & \textbf{$\text{RoBERTa}_{\text{BASE}}$} &  76.11\textit{(-06.76)} & 69.00\textit{(-13.87)} & 81.47\textit{(-01.40)} & 55.24\textit{(-27.63)}   & 77.58\textit{(-05.29)}\\
      & \textbf{$\text{RoBERTa}_{\text{LARGE}}$} & 83.80\textit{(-03.61)} & 74.24\textit{(-13.17)} & 86.60\textit{(-00.81)} & 60.84\textit{(-26.57)} & 85.12\textit{(-02.29)}\\

    \bottomrule
    \end{tabular}
        \captionsetup{font={small}} 
    \caption{Accuracy of PLMs in the zero-shot setting before and after hiding the information of each component on two datasets.}
    \label{tab:component_result}
\end{table*}

\subsection{Influence of Important Components}
\subsubsection{Experiment Set-up}
Due to the high performance of off-the-shelf PLMs, we are interested in the contributions of each component to infer shared properties in the zero-shot setting.
First, the information of each component is hidden through a certain strategy.
Specifically, for \textit{topic}, \textit{vehicle} and \textit{comparator}, we replace their tokens with a special token \texttt{[UNK]} which means unknown.
With regard to \textit{event}, we convert it into a suitable copula, such as ``am'' and ``is'', to ensure the integrity of syntax.
Furthermore, we also set up a baseline by randomly replacing a token with \texttt{[UNK]} in the context.
Examples corresponding to all settings are shown in Table \ref{tab:component_example} in the Appendix \ref{sec:experimental_details_appendix}.
We finally report the model accuracy and declined absolute accuracy after hiding the information of each component.

\subsubsection{Results}
The results in Table \ref{tab:component_result} show varying degrees of the decline of all settings.
If the model's performance decreases more, it means that the influence of the component is more significant than others.
Three major components (i.e., vehicle, topic and comparator) obtain higher declined absolute accuracy than random token, which demonstrates that the information of these simile components is more valuable than other words to infer the shared properties.
Among all the components, removing the \textit{comparator} may cause the most significant performance drop.
This result is mostly because PLMs cannot identify the sentence as a simile without an obvious indicator.
When it comes to the remaining 3 components, \textit{vehicle} contributes the most, followed by \textit{topic}.
Hence, we argue that it may be beneficial to explicitly leverage both the information of \textit{vehicle} and \textit{topic} to infer the properties.

\section{Enhancing PLMs with Knowledge} 

\begin{figure}[t]
    \centering
    \vspace{-0.3cm}
        \includegraphics[width=0.9\linewidth]{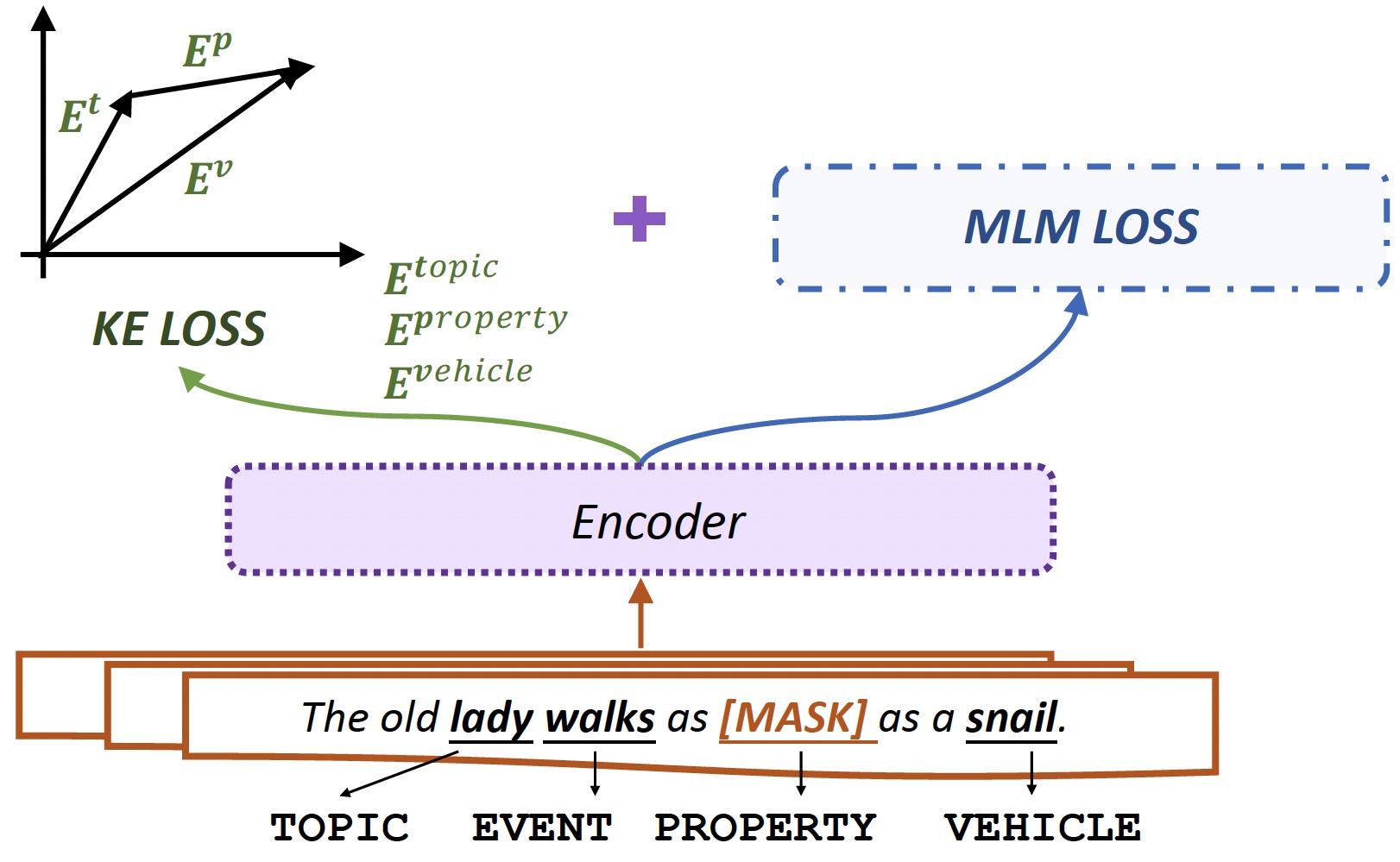} 
    \captionsetup{font={small}} 
    \caption{An overview of our objective function design}
    \label{fig:loss_framework}
\end{figure}

\subsection{Knowledge-enhanced Objective}
Benefiting from the result that topic and vehicle are the two most essential components for predicting the shared properties of similes, we catch an insight that \textit{property} can be seen as the \textbf{relation} between \textit{topic} and \textit{vehicle} following a set of knowledge embedding (KE) methods \cite{bordes2013translating,wang2014knowledge,ji2015knowledge}.

To integrate the insight mentioned above into our training procedure, we design an objective function as shown in Figure \ref{fig:loss_framework}.
Inspired by triplets representing the relational facts, we can also extract the topic, property, and vehicle from a simile as a triplet $(t,p,v)$.
The distance between topic and vehicle in the embedding space represents the plausibility of property.
The plausibility can be measured by scoring functions \cite{bordes2013translating,wang2014knowledge,ji2015knowledge}.
To this end, we follow the scoring function from TransE \cite{bordes2013translating} and define the following Mean Square Error (MSE) loss as our KE loss: 
\begin{align}
		&\mathcal{L}_\text{KE} = \text{MSE}(E^{t} + E^{p},E^{v})
\end{align} 
where $E^{t},E^{p},E^{v}$ are the representations of topic, property and vehicle encoded by PLMs.
We also try more advanced methods such as TransH \cite{wang2014knowledge} and TransD \cite{ji2015knowledge} for the knowledge embedding objective, and their results are presented in Table \ref{tab:trans_loss_result} in the Appendix \ref{sec:trans_loss_appendix}.

Finally, our training procedure is to optimize MLM loss and KE loss jointly:
\begin{align}
		&\mathcal{L}_\text{Ours} = \alpha \mathcal{L}_\text{KE} + \mathcal{L}_\text{MLM} 
\end{align} 
where $\alpha$ is a hyperparameter used to balance two objective functions. 

 
\subsection{Results}
\begin{table}[t]
    \small
    \centering
    \begin{tabular}{ccccc}
    \toprule
     \textbf{Datasets}  & \textbf{Models} & $\mathcal{L}_\text{MLM}$ & $\mathcal{L}_\text{Ours}$  & \textbf{Gain}   \\
        \midrule
        \multirow{4}{*}{\begin{tabular}[c]{@{}l@{}}\textbf{General}\\ \textbf{Corpus}\end{tabular}} & \textbf{$\text{BERT}_{\text{BASE}}$} & 67.74 & 69.25 & +1.51\\
        & \textbf{$\text{BERT}_{\text{LARGE}}$} & 73.85 & 74.07 & +0.22 \\
        & \textbf{$\text{RoBERTa}_{\text{BASE}}$} & 70.58 & 71.74 & +1.16 \\
        & \textbf{$\text{RoBERTa}_{\text{LARGE}}$} & 78.97 & 78.97 & +0.00\\
        \midrule
        \multirow{4}{*}{\textbf{Quizzes}} &  \textbf{$\text{BERT}_{\text{BASE}}$} & 82.05 & 82.94 & +0.89\\
        & \textbf{$\text{BERT}_{\text{LARGE}}$} & 84.58 & 85.94 & +1.36\\
        &\textbf{$\text{RoBERTa}_{\text{BASE}}$} & 84.69 & 84.89 & +0.20  \\
        &\textbf{$\text{RoBERTa}_{\text{LARGE}}$} & 88.97 & 89.40 & +0.43\\
        
    \bottomrule
    
    \end{tabular}
    \caption{Accuracy of PLMs using MLM and our objectives in our probing task.}
    \vspace{-0.2cm}
        \label{tab:loss_result}
\end{table}




Table \ref{tab:loss_result} presents the performance of the models fine-tuned with the MLM objective and our knowledge-enhanced objective on the two datasets, where the last column shows the performance gains brought by our improvement to the training objective.
Overall, each model trained with our knowledge-enhanced objective outperforms the one trained with the MLM objective on both datasets, demonstrating the effectiveness of our objective in the probing task.

For the Quizzes dataset, $\text{BERT}$ achieves more performance gains than $\text{RoBERTa}$ does, which is probably because $\text{RoBERTa}$ has better modeled the relationship among \textit{topic}, \textit{property} and \textit{vehicle} in the similes with simple syntactic structure during fine-tuning with the MLM objective.
For the General Corpus dataset, the BASE version of models tends to yield higher performance improvements, probably because the models with larger parameter sizes can better capture the relationship among simile components in the similes with more diverse contexts when fine-tuning with the MLM objective.


\subsection{Experiments with Downstream Tasks}
Similes generally transmit a positive or negative view due to the shared properties \cite{fishelov2007shall,li2012using,qadir2015learning}.
Taking the simile \textit{``the lawyer is like a shark''} as an example, the implicit shared property \textit{``aggressive''} between \textit{``lawyer''} and \textit{``shark''} indicates the negative polarity.
Therefore, we design a sentiment polarity downstream task to validate the improvement of our method to infer shared properties. 

\begin{table}[t]
    \small
    \centering
    \begin{tabular}{cccc}
    \toprule
     \textbf{Models} & \textbf{Original} & $\mathcal{L}_{\text{MLM}} $& $\mathcal{L}_{\text{Ours}}$    \\
        \midrule
        \textbf{$\text{BERT}_{\text{BASE}}$} & 84.96 & 85.45 & \textbf{85.63} \\
        \textbf{$\text{BERT}_{\text{LARGE}}$} & 86.02 & 86.65 & \textbf{86.95}\\
        \textbf{$\text{RoBERTa}_{\text{BASE}}$} & 88.51 & 88.61 & \textbf{89.51} \\
        \textbf{$\text{RoBERTa}_{\text{LARGE}}$} & 88.84 & 89.08 & \textbf{90.21} \\
    \bottomrule
    \end{tabular}
    \caption{ Accuracy of PLMs with three settings in the downstream task of sentiment classification.}
        \label{tab:sentiment_result}
\end{table}


Our experiments are based on the Amazon reviews dataset$\footnote{https://www.kaggle.com/bittlingmayer/amazonreviews}$ which provides reviews and their corresponding sentiment ratings.
Following \cite{mudinas2012combining,haque2018sentiment}, we first process the dataset into a binary sentiment classification task by defining the 1-star and 2-star ratings as negative, the 4-star, and 5-star ratings as positive, while excluding the 3-star neutral ratings.
To further address the label imbalance problem, we then sample the positive and negative reviews at 1:1.
The final dataset consists of 5,023 reviews and is split into a ratio of 6:2:2 for the train/dev/test set.


When performing the sentiment classification task, we only update the parameters of the multi-layer perceptron (MLP) classifiers on top of PLM's contextualized representation.
The parameters of PLM are fixed and from three settings: (1) zero-shot; (2) fine-tuned with the MLM objective in the probing task; (3) fine-tuned with the knowledge-enhanced objective in the probing task.
The results are shown in the Table \ref{tab:sentiment_result}.
First of all, fine-tuning with the MLM objective improves the performance of all models in the sentiment classification task, demonstrating that improving models' ability to infer the properties of similes can enhance models' understanding of the sentiment polarity.
Moreover, the performance is further improved by our knowledge-enhanced objective, especially for $\text{RoBERTa}$ whose main gains are mostly contributed by our additional knowledge embedding objective.
This indicates the effectiveness of our knowledge-enhanced objective in the downstream task of sentiment analysis.

\begin{figure}[t]
    \centering
    \vspace{-0.4cm}
        \includegraphics[width=0.8\linewidth]{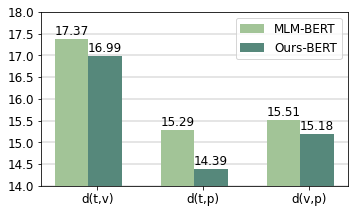} 
    \vspace{-0.08cm}
    \captionsetup{font={small}} 
    \caption{The average semantic distances between the representations of topic(t), property(p), and vehicle(v) in the last layer's hidden state given by $\text{BERT}_{\text{BASE}}$ with MLM and our objectives.}
        \label{fig:distance_result}
     \vspace{-0.4cm}
\end{figure}


\subsection{Analysis}

Furthermore, we investigate the mechanism of how knowledge-enhanced objective brings improvement.
We first calculate the L2 distance between the representations in the last hidden states of each pair of components.
The results are shown in Figure \ref{fig:distance_result}.
In all pairs, the distance given by our objective is generally shorter than MLM-BERT, which indicates that modeling the relationships among the three important components is efficient to enhance the model performance.



\begin{figure}[t]
    \centering
    \vspace{-1cm}
        \includegraphics[width=0.9\linewidth]{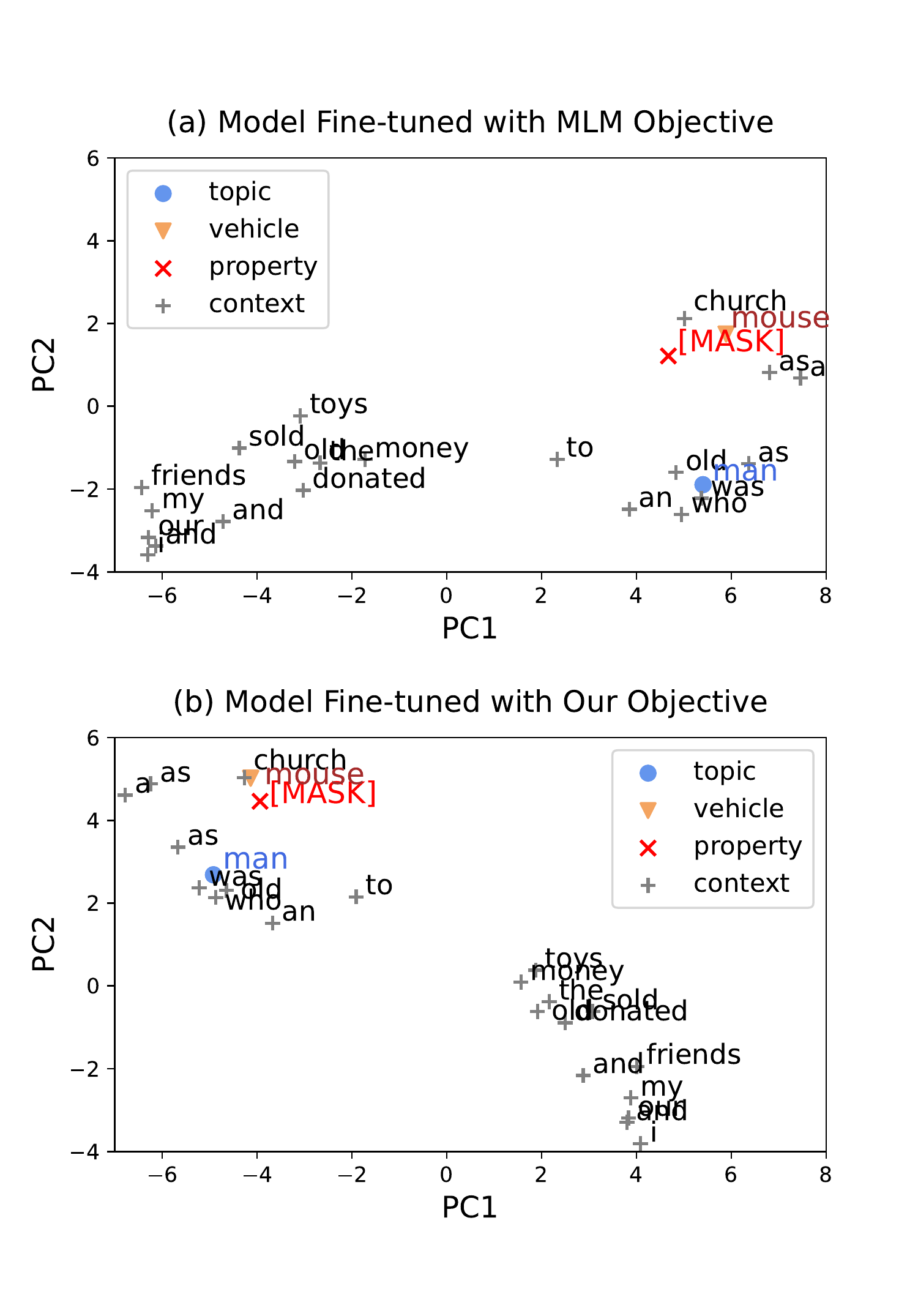} 
    \vspace{-0.6cm}
    \captionsetup{font={small}} 
    \caption{PCA representations of tokens in the last layer's hidden state given by $\text{BERT}_{\text{BASE}}$ with MLM and our objectives. }\label{fig:case_true}
\end{figure}

Specifically, we visualize the final layer representation of a simile into two-dimensional spaces via Principal Component Analysis (PCA) \cite{pearson1901liii} in Figure \ref{fig:case_true}.
In both MLM and our objective, the models are required to fill in the masked token in the same simile sentence. 
The model fine-tuned with the MLM objective predicts wrongly, while our fine-tuned model predicts correctly.
We find that our representations of the three components are closer to each other.


\section{Related Work}
\textbf{Simile Processing.}
Simile processing mainly involves 3 fields: simile detection, simile generation, and simile interpretation.
The bulk of work in similes mainly focuses on identifying similes and their components \cite{niculae2013comparison,niculae2014brighter,liu2018neural,zeng2020neural}. 
Recent years have witnessed a growth of work to transfer literal sentences to similes \cite{zhang2020writing,chakrabarty2020generating}. 
\cite{chakrabarty2021figurative} study the ability of PLMs to recognize textual entailment related to similes.
With regard to simile interpretation, \cite{qadir2016automatically,xiao2016meta4meaning,bar2020automatic,zheng2019love} rank the properties by the statistical co-occurrence and embedding similarities with other simile components.
\cite{chakrabarty2021s} interpret similes by choosing or generating continuation for narratives via PLMs.
Different from these works, we investigate the ability of PLMs to infer shared properties of similes.

\textbf{Probing Tasks for PLMs.}
Many studies investigate whether PLMs encode knowledge in their contextual representations by designing probing tasks.
Early studies mainly focus on the linguistic knowledge captured by PLMs \cite{liu2019linguistic,tenney2019you}.
 \cite{petroni2019language} first propose a word prediction task to probe factual knowledge stored in PLMs.
Similar methods are utilized to evaluate various commonsense knowledge, such as symbolic reasoning ability \cite{talmor2020olmpics,zhou2020evaluating}, numerical commonsense knowledge \cite{lin2020birds}, properties associated with concepts \cite{weir2020probing}.
To our best knowledge, we are the first work to investigate the ability of PLMs to interpret similes by proposing a simile property probing task.

\textbf{Enhance PLMs via Knowledge Regularization.}
Recently, many researchers integrate external knowledge with PLMs by complementing the MLM objective with an auxiliary knowledge-based objective.
For example, there are works that introduce span-boundary objective for span-level prediction \cite{joshi2020spanbert}, copy-based training objective for mention reference prediction \cite{ye2020coreferential}, knowledge embedding objective for factual knowledge \cite{wang2021kepler} and arithmetic relationships of linguistic units for universal language representation \cite{li2021pre}.
Different from these works, we incorporate simile knowledge with the training objective by modeling the relationship between the salient components of similes.

\section{Conclusion}
In this work, we are the first to investigate the ability of PLMs in simile interpretation via a proposed novel simile property probing task.
We construct two multi-choice probing datasets covering two data sources.
By conducting a series of empirical experiments, we prove that PLMs exhibit the ability to infer simile properties in the pre-training stage and further induce more related knowledge during the fine-tuning stage, but there is still a gap between PLMs and humans in this task.
Furthermore, we propose a knowledge-enhanced training objective to bridge the gap, which shows effectiveness in the probing task and the downstream task of sentiment classification.
In future work, we are interested in exploring the interpretation of more sophisticated figurative language, such as metaphor or analogy.

\section*{Acknowledgements}
We would like to thank anonymous reviews for their helpful comments and suggestions.
Also, thanks to Jingping Liu, Leyang Cui for their insightful feedback that helped improve the paper.
We also thank Botian Jiang, Shuang Li for supporting our data collection.
This research was supported by the National Key Research and Development Project
(No. 2020AAA0109302), National Natural Science Foundation of China (No. 62072323),  Shanghai Science and Technology Innovation
Action Plan (No. 19511120400), Shanghai Municipal Science
and Technology Major Project (No. 2021SHZDZX0103).

\section*{Ethical Consideration}
We provide details of our work to address potential ethical concerns.
In our work, we propose a simile property probing task and construct probing datasets from both general textual corpora and human-designed questions.
First of all, all the data sources used in the data collection process are publicly available. 
Specifically, we follow the robots.txt$\footnote{https://quizizz.com/robots.txt}$ to respect the copyright when we collect similes from the learning platform Quizizz (Sec. \ref{sec:data_sources}).
Moreover, there are three steps involving human annotation to ensure the quality of the datasets: simile and simile components recognition (Sec. \ref{sec:data_sources}), human confirmation for distractors (Sec. \ref{sec:distractor_design}), and human performance (Sec. \ref{sec:ability_to_infer}).
To ensure the quality of annotation, all the annotators do not participate in our probing data collection, and they always label a small set of 50 examples to reach an agreement on the labeling criteria before the formal labeling.
We protect the privacy rights of annotators and pay them above the local minimum wage.



\bibliography{anthology, custom}

\begin{thebibliography}{52}
\expandafter\ifx\csname natexlab\endcsname\relax\def\natexlab#1{#1}\fi

\bibitem[{Bar et~al.(2020)Bar, Dershowitz, and Dankin}]{bar2020automatic}
Kfir Bar, Nachum Dershowitz, and Lena Dankin. 2020.
\newblock Automatic metaphor interpretation using word embeddings.
\newblock \emph{arXiv preprint arXiv:2010.02665}.

\bibitem[{Beardsley(1981)}]{beardsley1981aesthetics}
Monroe~C Beardsley. 1981.
\newblock \emph{Aesthetics, problems in the philosophy of criticism}.
\newblock Hackett Publishing.

\bibitem[{Bordes et~al.(2013)Bordes, Usunier, Garcia-Duran, Weston, and
  Yakhnenko}]{bordes2013translating}
Antoine Bordes, Nicolas Usunier, Alberto Garcia-Duran, Jason Weston, and Oksana
  Yakhnenko. 2013.
\newblock Translating embeddings for modeling multi-relational data.
\newblock \emph{Advances in neural information processing systems}, 26.

\bibitem[{Bosselut et~al.(2019)Bosselut, Rashkin, Sap, Malaviya, Celikyilmaz,
  and Choi}]{bosselut2019comet}
Antoine Bosselut, Hannah Rashkin, Maarten Sap, Chaitanya Malaviya, Asli
  Celikyilmaz, and Yejin Choi. 2019.
\newblock Comet: Commonsense transformers for automatic knowledge graph
  construction.
\newblock \emph{arXiv preprint arXiv:1906.05317}.

\bibitem[{Chakrabarty et~al.(2021{\natexlab{a}})Chakrabarty, Choi, and
  Shwartz}]{chakrabarty2021s}
Tuhin Chakrabarty, Yejin Choi, and Vered Shwartz. 2021{\natexlab{a}}.
\newblock It's not rocket science: Interpreting figurative language in
  narratives.
\newblock \emph{arXiv preprint arXiv:2109.00087}.

\bibitem[{Chakrabarty et~al.(2021{\natexlab{b}})Chakrabarty, Ghosh, Poliak, and
  Muresan}]{chakrabarty2021figurative}
Tuhin Chakrabarty, Debanjan Ghosh, Adam Poliak, and Smaranda Muresan.
  2021{\natexlab{b}}.
\newblock Figurative language in recognizing textual entailment.
\newblock \emph{arXiv preprint arXiv:2106.01195}.

\bibitem[{Chakrabarty et~al.(2020{\natexlab{a}})Chakrabarty, Muresan, and
  Peng}]{chakrabarty-etal-2020-generating}
Tuhin Chakrabarty, Smaranda Muresan, and Nanyun Peng. 2020{\natexlab{a}}.
\newblock \href {https://doi.org/10.18653/v1/2020.emnlp-main.524} {Generating
  similes effortlessly like a pro: A style transfer approach for simile
  generation}.
\newblock In \emph{Proceedings of the 2020 Conference on Empirical Methods in
  Natural Language Processing (EMNLP)}, pages 6455--6469, Online. Association
  for Computational Linguistics.

\bibitem[{Chakrabarty et~al.(2020{\natexlab{b}})Chakrabarty, Muresan, and
  Peng}]{chakrabarty2020generating}
Tuhin Chakrabarty, Smaranda Muresan, and Nanyun Peng. 2020{\natexlab{b}}.
\newblock Generating similes< effortlessly> like a pro: A style transfer
  approach for simile generation.
\newblock \emph{arXiv preprint arXiv:2009.08942}.

\bibitem[{Cui et~al.(2021)Cui, Cheng, Wu, and Zhang}]{cui2021commonsense}
Leyang Cui, Sijie Cheng, Yu~Wu, and Yue Zhang. 2021.
\newblock On commonsense cues in bert for solving commonsense tasks.
\newblock In \emph{Findings of the Association for Computational Linguistics:
  ACL-IJCNLP 2021}, pages 683--693.

\bibitem[{Devlin et~al.(2018)Devlin, Chang, Lee, and
  Toutanova}]{devlin2018bert}
Jacob Devlin, Ming-Wei Chang, Kenton Lee, and Kristina Toutanova. 2018.
\newblock Bert: Pre-training of deep bidirectional transformers for language
  understanding.
\newblock \emph{arXiv preprint arXiv:1810.04805}.

\bibitem[{Fishelov(2007)}]{fishelov2007shall}
David Fishelov. 2007.
\newblock Shall i compare thee? simile understanding and semantic categories.

\bibitem[{Fleiss(1971)}]{fleiss1971measuring}
Joseph~L Fleiss. 1971.
\newblock Measuring nominal scale agreement among many raters.
\newblock \emph{Psychological bulletin}, 76(5):378.

\bibitem[{Gerlach and Font-Clos(2020)}]{gerlach2020standardized}
Martin Gerlach and Francesc Font-Clos. 2020.
\newblock A standardized project gutenberg corpus for statistical analysis of
  natural language and quantitative linguistics.
\newblock \emph{Entropy}, 22(1):126.

\bibitem[{Goldberg(2019)}]{goldberg2019assessing}
Yoav Goldberg. 2019.
\newblock Assessing bert's syntactic abilities.
\newblock \emph{arXiv preprint arXiv:1901.05287}.

\bibitem[{Haladyna et~al.(2002)Haladyna, Downing, and
  Rodriguez}]{haladyna2002review}
Thomas~M Haladyna, Steven~M Downing, and Michael~C Rodriguez. 2002.
\newblock A review of multiple-choice item-writing guidelines for classroom
  assessment.
\newblock \emph{Applied measurement in education}, 15(3):309--333.

\bibitem[{Hanks(2005)}]{hanks2005similes}
Patrick Hanks. 2005.
\newblock Similes and sets: The english preposition like.
\newblock \emph{Languages and Linguistics: Festschrift for Fr. Cermak. Charles
  University, Prague}.

\bibitem[{Hanks(2013)}]{hanks2013lexical}
Patrick Hanks. 2013.
\newblock \emph{Lexical analysis: Norms and exploitations}.
\newblock Mit Press.

\bibitem[{Haque et~al.(2018)Haque, Saber, and Shah}]{haque2018sentiment}
Tanjim~Ul Haque, Nudrat~Nawal Saber, and Faisal~Muhammad Shah. 2018.
\newblock Sentiment analysis on large scale amazon product reviews.
\newblock In \emph{2018 IEEE international conference on innovative research
  and development (ICIRD)}, pages 1--6. IEEE.

\bibitem[{Ji et~al.(2015)Ji, He, Xu, Liu, and Zhao}]{ji2015knowledge}
Guoliang Ji, Shizhu He, Liheng Xu, Kang Liu, and Jun Zhao. 2015.
\newblock Knowledge graph embedding via dynamic mapping matrix.
\newblock In \emph{Proceedings of the 53rd Annual Meeting of the Association
  for Computational Linguistics and the 7th International Joint Conference on
  Natural Language Processing (Volume 1: Long Papers)}, pages 687--696.

\bibitem[{Joshi et~al.(2020)Joshi, Chen, Liu, Weld, Zettlemoyer, and
  Levy}]{joshi2020spanbert}
Mandar Joshi, Danqi Chen, Yinhan Liu, Daniel~S Weld, Luke Zettlemoyer, and Omer
  Levy. 2020.
\newblock Spanbert: Improving pre-training by representing and predicting
  spans.
\newblock \emph{Transactions of the Association for Computational Linguistics},
  8:64--77.

\bibitem[{Lacroix et~al.(2005)Lacroix, Murre, and
  Postma}]{lacroix2005interpretive}
Joyca~PW Lacroix, Jaap~MJ Murre, and Eric~O Postma. 2005.
\newblock Interpretive diversity as a source of metaphor-simile distinction.
\newblock In \emph{Proceedings of the Annual Meeting of the Cognitive Science
  Society}, volume~27.

\bibitem[{Li et~al.(2012)Li, Kuang, Zhang, Chen, and Tang}]{li2012using}
Bin Li, Haibo Kuang, Yingjie Zhang, Jiajun Chen, and Xuri Tang. 2012.
\newblock Using similes to extract basic sentiments across languages.
\newblock In \emph{International Conference on Web Information Systems and
  Mining}, pages 536--542. Springer.

\bibitem[{Li and Zhao(2021)}]{li2021pre}
Yian Li and Hai Zhao. 2021.
\newblock Pre-training universal language representation.
\newblock \emph{arXiv preprint arXiv:2105.14478}.

\bibitem[{Lin et~al.(2019)Lin, Chen, Chen, and Ren}]{lin2019kagnet}
Bill~Yuchen Lin, Xinyue Chen, Jamin Chen, and Xiang Ren. 2019.
\newblock Kagnet: Knowledge-aware graph networks for commonsense reasoning.
\newblock \emph{arXiv preprint arXiv:1909.02151}.

\bibitem[{Lin et~al.(2020)Lin, Lee, Khanna, and Ren}]{lin2020birds}
Bill~Yuchen Lin, Seyeon Lee, Rahul Khanna, and Xiang Ren. 2020.
\newblock Birds have four legs?! numersense: Probing numerical commonsense
  knowledge of pre-trained language models.
\newblock \emph{arXiv preprint arXiv:2005.00683}.

\bibitem[{Liu and Singh(2004)}]{liu2004conceptnet}
Hugo Liu and Push Singh. 2004.
\newblock Conceptnet—a practical commonsense reasoning tool-kit.
\newblock \emph{BT technology journal}, 22(4):211--226.

\bibitem[{Liu et~al.(2018)Liu, Hu, Song, Fu, Liu, and Hu}]{liu2018neural}
Lizhen Liu, Xiao Hu, Wei Song, Ruiji Fu, Ting Liu, and Guoping Hu. 2018.
\newblock Neural multitask learning for simile recognition.
\newblock In \emph{Proceedings of the 2018 Conference on Empirical Methods in
  Natural Language Processing}, pages 1543--1553.

\bibitem[{Liu et~al.(2019{\natexlab{a}})Liu, Gardner, Belinkov, Peters, and
  Smith}]{liu2019linguistic}
Nelson~F Liu, Matt Gardner, Yonatan Belinkov, Matthew~E Peters, and Noah~A
  Smith. 2019{\natexlab{a}}.
\newblock Linguistic knowledge and transferability of contextual
  representations.
\newblock \emph{arXiv preprint arXiv:1903.08855}.

\bibitem[{Liu et~al.(2019{\natexlab{b}})Liu, Ott, Goyal, Du, Joshi, Chen, Levy,
  Lewis, Zettlemoyer, and Stoyanov}]{liu2019roberta}
Yinhan Liu, Myle Ott, Naman Goyal, Jingfei Du, Mandar Joshi, Danqi Chen, Omer
  Levy, Mike Lewis, Luke Zettlemoyer, and Veselin Stoyanov. 2019{\natexlab{b}}.
\newblock Roberta: A robustly optimized bert pretraining approach.
\newblock \emph{arXiv preprint arXiv:1907.11692}.

\bibitem[{Miller(1995)}]{miller1995wordnet}
George~A Miller. 1995.
\newblock Wordnet: a lexical database for english.
\newblock \emph{Communications of the ACM}, 38(11):39--41.

\bibitem[{Mudinas et~al.(2012)Mudinas, Zhang, and
  Levene}]{mudinas2012combining}
Andrius Mudinas, Dell Zhang, and Mark Levene. 2012.
\newblock Combining lexicon and learning based approaches for concept-level
  sentiment analysis.
\newblock In \emph{Proceedings of the first international workshop on issues of
  sentiment discovery and opinion mining}, pages 1--8.

\bibitem[{Niculae(2013)}]{niculae2013comparison}
Vlad Niculae. 2013.
\newblock Comparison pattern matching and creative simile recognition.
\newblock In \emph{Proceedings of the Joint Symposium on Semantic Processing.
  Textual Inference and Structures in Corpora}, pages 110--114.

\bibitem[{Niculae and Danescu-Niculescu-Mizil(2014)}]{niculae2014brighter}
Vlad Niculae and Cristian Danescu-Niculescu-Mizil. 2014.
\newblock Brighter than gold: Figurative language in user generated
  comparisons.
\newblock In \emph{Proceedings of the 2014 conference on empirical methods in
  natural language processing (EMNLP)}, pages 2008--2018.

\bibitem[{Niculae and Yaneva(2013)}]{niculae2013computational}
Vlad Niculae and Victoria Yaneva. 2013.
\newblock Computational considerations of comparisons and similes.
\newblock In \emph{51st Annual Meeting of the Association for Computational
  Linguistics Proceedings of the Student Research Workshop}, pages 89--95.

\bibitem[{Paul(1970)}]{paul1970figurative}
Anthony~M Paul. 1970.
\newblock Figurative language.
\newblock \emph{Philosophy \& Rhetoric}, pages 225--248.

\bibitem[{Pearson(1901)}]{pearson1901liii}
Karl Pearson. 1901.
\newblock Liii. on lines and planes of closest fit to systems of points in
  space.
\newblock \emph{The London, Edinburgh, and Dublin philosophical magazine and
  journal of science}, 2(11):559--572.

\bibitem[{Petroni et~al.(2019)Petroni, Rockt{\"a}schel, Lewis, Bakhtin, Wu,
  Miller, and Riedel}]{petroni2019language}
Fabio Petroni, Tim Rockt{\"a}schel, Patrick Lewis, Anton Bakhtin, Yuxiang Wu,
  Alexander~H Miller, and Sebastian Riedel. 2019.
\newblock Language models as knowledge bases?
\newblock \emph{arXiv preprint arXiv:1909.01066}.

\bibitem[{Qadir et~al.(2015)Qadir, Riloff, and Walker}]{qadir2015learning}
Ashequl Qadir, Ellen Riloff, and Marilyn Walker. 2015.
\newblock Learning to recognize affective polarity in similes.
\newblock In \emph{Proceedings of the 2015 Conference on Empirical Methods in
  Natural Language Processing}, pages 190--200.

\bibitem[{Qadir et~al.(2016)Qadir, Riloff, and Walker}]{qadir2016automatically}
Ashequl Qadir, Ellen Riloff, and Marilyn Walker. 2016.
\newblock Automatically inferring implicit properties in similes.
\newblock In \emph{Proceedings of the 2016 Conference of the North American
  Chapter of the Association for Computational Linguistics: Human Language
  Technologies}, pages 1223--1232.

\bibitem[{Ren and Zhu(2020)}]{ren2020knowledge}
Siyu Ren and Kenny~Q Zhu. 2020.
\newblock Knowledge-driven distractor generation for cloze-style multiple
  choice questions.
\newblock \emph{arXiv preprint arXiv:2004.09853}.

\bibitem[{Talmor et~al.(2020)Talmor, Elazar, Goldberg, and
  Berant}]{talmor2020olmpics}
Alon Talmor, Yanai Elazar, Yoav Goldberg, and Jonathan Berant. 2020.
\newblock olmpics-on what language model pre-training captures.
\newblock \emph{Transactions of the Association for Computational Linguistics},
  8:743--758.

\bibitem[{Tenney et~al.(2019)Tenney, Xia, Chen, Wang, Poliak, McCoy, Kim,
  Van~Durme, Bowman, Das et~al.}]{tenney2019you}
Ian Tenney, Patrick Xia, Berlin Chen, Alex Wang, Adam Poliak, R~Thomas McCoy,
  Najoung Kim, Benjamin Van~Durme, Samuel~R Bowman, Dipanjan Das, et~al. 2019.
\newblock What do you learn from context? probing for sentence structure in
  contextualized word representations.
\newblock \emph{arXiv preprint arXiv:1905.06316}.

\bibitem[{Veale and Hao(2007)}]{veale2007learning}
Tony Veale and Yanfen Hao. 2007.
\newblock Learning to understand figurative language: From similes to metaphors
  to irony.
\newblock In \emph{Proceedings of the annual meeting of the cognitive science
  society}, volume~29.

\bibitem[{Wang et~al.(2021)Wang, Gao, Zhu, Zhang, Liu, Li, and
  Tang}]{wang2021kepler}
Xiaozhi Wang, Tianyu Gao, Zhaocheng Zhu, Zhengyan Zhang, Zhiyuan Liu, Juanzi
  Li, and Jian Tang. 2021.
\newblock Kepler: A unified model for knowledge embedding and pre-trained
  language representation.
\newblock \emph{Transactions of the Association for Computational Linguistics},
  9:176--194.

\bibitem[{Wang et~al.(2014)Wang, Zhang, Feng, and Chen}]{wang2014knowledge}
Zhen Wang, Jianwen Zhang, Jianlin Feng, and Zheng Chen. 2014.
\newblock Knowledge graph embedding by translating on hyperplanes.
\newblock In \emph{Proceedings of the AAAI Conference on Artificial
  Intelligence}, volume~28.

\bibitem[{Weir et~al.(2020)Weir, Poliak, and Van~Durme}]{weir2020probing}
Nathaniel Weir, Adam Poliak, and Benjamin Van~Durme. 2020.
\newblock Probing neural language models for human tacit assumptions.
\newblock \emph{arXiv preprint arXiv:2004.04877}.

\bibitem[{Xiao et~al.(2016)Xiao, Alnajjar, Granroth-Wilding, Agres, Toivonen
  et~al.}]{xiao2016meta4meaning}
Ping Xiao, Khalid Alnajjar, Mark Granroth-Wilding, Kat Agres, Hannu Toivonen,
  et~al. 2016.
\newblock Meta4meaning: Automatic metaphor interpretation using corpus-derived
  word associations.
\newblock In \emph{Proceedings of the Seventh International Conference on
  Computational Creativity}. Sony CSL Paris.

\bibitem[{Ye et~al.(2020)Ye, Lin, Du, Liu, Li, Sun, and
  Liu}]{ye2020coreferential}
Deming Ye, Yankai Lin, Jiaju Du, Zhenghao Liu, Peng Li, Maosong Sun, and
  Zhiyuan Liu. 2020.
\newblock Coreferential reasoning learning for language representation.
\newblock \emph{arXiv preprint arXiv:2004.06870}.

\bibitem[{Zeng et~al.(2020)Zeng, Song, Su, Xie, Song, and Luo}]{zeng2020neural}
Jiali Zeng, Linfeng Song, Jinsong Su, Jun Xie, Wei Song, and Jiebo Luo. 2020.
\newblock Neural simile recognition with cyclic multitask learning and local
  attention.
\newblock In \emph{Proceedings of the AAAI Conference on Artificial
  Intelligence}, volume~34, pages 9515--9522.

\bibitem[{Zhang et~al.(2020)Zhang, Cui, Xia, Guo, Li, Wei, and
  Cui}]{zhang2020writing}
Jiayi Zhang, Zhi Cui, Xiaoqiang Xia, Yalong Guo, Yanran Li, Chen Wei, and
  Jianwei Cui. 2020.
\newblock Writing polishment with simile: Task, dataset and a neural approach.
\newblock \emph{arXiv preprint arXiv:2012.08117}.

\bibitem[{Zheng et~al.(2019)Zheng, Song, Hu, Fu, and Zhou}]{zheng2019love}
Danning Zheng, Ruihua Song, Tianran Hu, Hao Fu, and Jin Zhou. 2019.
\newblock “love is as complex as math”: Metaphor generation system for
  social chatbot.
\newblock In \emph{Workshop on Chinese Lexical Semantics}, pages 337--347.
  Springer.

\bibitem[{Zhou et~al.(2020)Zhou, Zhang, Cui, and Huang}]{zhou2020evaluating}
Xuhui Zhou, Yue Zhang, Leyang Cui, and Dandan Huang. 2020.
\newblock Evaluating commonsense in pre-trained language models.
\newblock In \emph{Proceedings of the AAAI Conference on Artificial
  Intelligence}, volume~34, pages 9733--9740.

\end{thebibliography}
\bibliographystyle{acl_natbib}

\clearpage

\appendix
\label{sec:appendix}

\section{Additional Experimental Results} \label{sec:additional_experimental_results_appendix}
\subsection{Performance on Different Categories} \label{sec:performance_on_different_categories_appendix}

\begin{figure}[h]
    \centering
        \includegraphics[width=1\linewidth]{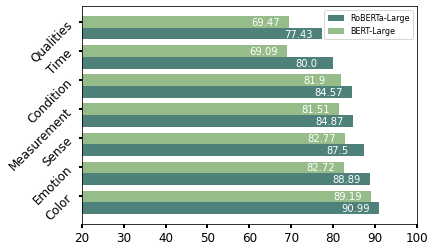} 
    \captionsetup{font={small}} 
    \caption{The average accuracy for each category in the zero-shot setting. We select the strongest version from each group of models. }\label{fig:acc_category}
\end{figure}

We investigate whether PLMs are better at inferring the properties of certain categories.
Figure \ref{fig:acc_category} presents the performance of the strongest version from each group of models for each category in the zero-shot setting.
We found that models perform significantly well when inferring the color, which is probably because each object often has a specific color which in many cases can be inferred without context.
However, when it comes to the properties requiring an understanding of the context, such as the personality and qualities \textit{(intelligent, brave)}, temporal properties \textit{(ancient, swift)} and short-term state \textit{(busy, safe)}, models tend to have relatively lower accuracy.


\subsection{Comparison of Knowledge Embedding Methods} \label{sec:trans_loss_appendix}

We also exploit the effects of different knowledge embedding methods when designing our knowledge-enhanced objective.
Table \ref{tab:trans_loss_result} shows the performance given by the objectives applying different knowledge embedding methods. 
First of all, complementing the MLM objective with our knowledge embedding methods generally improves the performance, demonstrating the effectiveness of our approach to enhancing PLMs with simile knowledge.
Moreover, following the scoring function from TransE \cite{bordes2013translating} brings the best result in most cases, which indicates that the knowledge embedding methods of simple design are sufficient to incorporate simile knowledge into PLMs in our objective design.

\begin{table}[h]
    \scriptsize
    \centering
    \begin{tabular}{cccccc}
    \toprule
     \textbf{Datasets}  & \textbf{Models} & $\mathcal{L}_\text{MLM}$ & $\mathcal{L}_\text{Ours}$  & $\mathcal{L}_\text{TransH}$ & $\mathcal{L}_\text{TransD}$    \\
        \midrule
        \multirow{4}{*}{\begin{tabular}[c]{@{}l@{}}\textbf{General}\\ \textbf{Corpus}\end{tabular}} & \textbf{$\text{BERT}_{\text{BASE}}$} & 67.74 & 69.25 & \textbf{69.72} & 68.38  \\
        & \textbf{$\text{BERT}_{\text{LARGE}}$} & 73.85 & 74.07 &  \textbf{74.33} & 73.85 \\
        & \textbf{$\text{RoBERTa}_{\text{BASE}}$} & 70.58 & \textbf{71.74} & 71.18 & 70.97 \\
        & \textbf{$\text{RoBERTa}_{\text{LARGE}}$} & 78.97 & 78.97 & 78.97 & 78.97\\
        \midrule
        \multirow{4}{*}{\textbf{Quizzes}} &  \textbf{$\text{BERT}_{\text{BASE}}$} & 82.05 & \textbf{82.94} & 82.25 & 82.05\\
        & \textbf{$\text{BERT}_{\text{LARGE}}$} & 84.58 & \textbf{85.94 }& 85.24 & 84.69 \\
        &\textbf{$\text{RoBERTa}_{\text{BASE}}$} & 84.69 & \textbf{84.89} & 84.81 & 84.81  \\
        &\textbf{$\text{RoBERTa}_{\text{LARGE}}$} & 88.97 & \textbf{89.40}& 89.32 & 88.96\\
        
    \bottomrule
    \end{tabular}
    \caption{ Comparison of different knowledge embedding methods when designing the knowledge-enhanced objective in our probing task.}
        \label{tab:trans_loss_result}
\end{table}

\section{Experimental Details} \label{sec:experimental_details_appendix}
We introduce details about the implementation of our experiments.
The implementations of all the PLMs in our paper are based on the HuggingFace Transformers$\footnote{https://github.com/huggingface/transformers/}$.
During fine-tuning for the probing task, the experiments are run with batch sizes in \{8, 16\}, $\alpha$ in \{3, 5, 10\}, a max sequence length of 128, and a learning rate of 1e-5 for 10 epochs.
For each model, we use the same hyper-parameters when applying different training objectives.
During fine-tuning for the sentiment analysis task, we only update the parameters of the multi-layer perceptron (MLP) classifiers on top of PLM's contextualized representation.
We set the learning rate in \{2e-5, 3e-5, 4e-5\}, batch size of 32, max sequence length of 128 and train for 200 epochs.
Additionally, we present examples of the experimental setup for evaluating the influence of important components in Table \ref{tab:component_example}.

\begin{table}[h] 
\scriptsize
    \centering
        \begin{tabular}{cccc}
        \toprule  
        \textbf{Component} & \textbf{Sentence Example} \\ 
        \midrule  
        \textbf{Original} & \makecell[c]{Johan runs as \texttt{[MASK]} as a deer to the toilet \\ after he had some spicy gravy  . } \\
        \hline
        \textbf{Topic} &  \makecell[c]{\texttt{[UNK]} runs as \texttt{[MASK]}  as a deer to the toilet \\ after he had some spicy gravy .}   \\
        \hline
        \textbf{Vehicle} & \makecell[c]{Johan runs as   \texttt{[MASK]}  as \texttt{[UNK]} to the toilet \\ after he had some spicy gravy .  }\\
        \hline
        \textbf{Event} & \makecell[c]{Johan is as \texttt{[MASK]}  as a deer to the toilet \\ after he had some spicy gravy .   }\\
        \hline
        \textbf{Comparator} & \makecell[c]{Johan runs \texttt{[UNK]} \texttt{[MASK]}   \texttt{[UNK]} a deer to the toilet \\ after he had some spicy gravy . }  \\
        \hline
        \textbf{Random} & \makecell[c]{Johan runs as  \texttt{[MASK]} as a deer \texttt{[UNK]} the toilet \\ after he had some spicy gravy . } \\
        \bottomrule 
        \end{tabular}
        \captionsetup{font={small}} 
        \caption{Examples of experiment set-up for evaluating the influence of important components.}
    \label{tab:component_example}
\end{table}

\section{Dataset Description} \label{sec:dataset_description_appendix}
We introduce details about our classification of the categories of properties.
We ask two annotators to label the category of each property in the given sentence and ensure that they agree on the questions that they gave completely different annotation results.
Table \ref{tab:adj_example} shows the percentage and five examples for each category (possibly more than one category per property).
In particular, properties in \textit{Quialities} describe the long-term feature of a material or a person's character, while properties in \textit{Condition} depict a short-term state.
Table \ref{tab:category_statistics} presents the percentage and examples for our simile probes of different categories.

\begin{table}[] 
    \scriptsize
    \centering
        \begin{tabular}{cccc}
        \toprule  
       \textbf{Category} & \textbf{Property Example } & \% \\ 
        \midrule  
      \textbf{Qualities }& strong, weak, cruel, intelligent, brave & 27.78 \\
        \hline
   \textbf{Condition} & bad, busy, idle, safe, vain & 22.28\\
        \hline
       \textbf{Sense} & cold, warm, bitter, soft, loud & 17.20\\
        \hline
      \textbf{Measurement} & big, scarce, numerous, tall, tiny & 14.16\\
        \hline
        \textbf{Color }& red, black, green, white, blue & 06.75 \\
        \hline
      \textbf{Time }& ancient, new, swift, slow, regular & 06.57\\
        \hline
      \textbf{Emotion} & excited, angry, sad, mad, nervous & 05.26 \\
        
        \bottomrule 
        \end{tabular}
        \captionsetup{font={small}} 
        \caption{Percentage and examples of each category of properties in constructed simile property probing datasets.}
    \label{tab:adj_example}
\end{table}

\end{document}